\documentclass[10pt,journal,compsoc]{IEEEtran}
\pdfoutput=1
\ifCLASSOPTIONcompsoc
  \usepackage[nocompress]{cite}
\else
  \usepackage{cite}
\fi
\ifCLASSINFOpdf
  \usepackage[pdftex]{graphicx}
  \DeclareGraphicsExtensions{.pdf,.jpeg,.png}
\else
  \usepackage[dvips]{graphicx}
  \DeclareGraphicsExtensions{.eps}
\fi
\usepackage{amsmath,amssymb,amsfonts,dsfont,pifont,bm,bbm,mathrsfs,mathtools}
\usepackage{algorithm,algpseudocode,listings}
\usepackage{booktabs,multirow,adjustbox,diagbox}
\usepackage{xspace}
\usepackage{float,capt-of}
\usepackage[table]{xcolor}
\usepackage{url}
\usepackage{makecell}
\usepackage[pagebackref=false,breaklinks=true,colorlinks=true,citecolor=blue,bookmarks=false]{hyperref}
\usepackage[capitalize]{cleveref}  %
\usepackage{tabularray}
\usepackage{subcaption}
\usepackage{fourier}
\usepackage{graphicx} 
\usepackage{grffile} 
\crefname{section}{Sec.}{Secs.}
\Crefname{section}{Section}{Sections}
\crefname{table}{Tab.}{Tabs.}
\Crefname{table}{Table}{Tables}
\crefname{figure}{Fig.}{Figs.}
\Crefname{figure}{Figure}{Figures}
\crefname{equation}{Eq.}{Eqs.}
\Crefname{equation}{Equation}{Equations}
\usepackage{tikz}
\usetikzlibrary{tikzmark, calc}
\tikzset{
    double color fill/.code 2 args={
        \pgfdeclareverticalshading[%
            tikz@axis@top,tikz@axis@middle,tikz@axis@bottom%
        ]{diagonalfill}{100bp}{%
            color(0bp)=(tikz@axis@bottom);
            color(50bp)=(tikz@axis@bottom);
            color(50bp)=(tikz@axis@middle);
            color(50bp)=(tikz@axis@top);
            color(100bp)=(tikz@axis@top)
        }
        \tikzset{shade, left color=#1, right color=#2, shading=diagonalfill}
    }
}
\tikzset{%
    diagonal fill/.style 2 args={%
        double color fill={#1}{#2},
        shading angle=45,
        opacity=0.8},
    filling 1/.style={%
        shade,
        shading=myshade1,
        shading angle=0,
        opacity=0.5},
    filling 2/.style={%
        shade,
        shading=myshade2,
        shading angle=0,
        opacity=0.5},
    filling 3/.style={%
        shade,
        shading=myshade3,
        shading angle=0,
        opacity=0.5}
   }
\pgfdeclarehorizontalshading{myshade1}{100bp}{%
      color(0bp)=(orange!30);
      color(50bp)=(orange!30);
      color(50bp)=(yellow!30);
      color(100bp)=(yellow!30)}

\pgfdeclarehorizontalshading{myshade2}{100bp}{%
      color(0bp)=(orange!30);
      color(50bp)=(orange!30);
      color(50bp)=(red!30);
      color(100bp)=(red!30)}
\pgfdeclarehorizontalshading{myshade3}{100bp}{%
      color(0bp)=(orange!30);
      color(50bp)=(orange!30);
      color(50bp)=(cyan!30);
      color(100bp)=(cyan!30)}
\newcounter{BGnum}
\definecolor{color1}{rgb}{0.87, 0.894, 0.768}
\definecolor{color2}{rgb}{0.90,0.773,0.773}%
\definecolor{color3}{rgb}{0.773, 0.824, 0.886}
\definecolor{color4}{rgb}{0.95, 0.83, 0.71}

\usepackage{dcolumn, array, booktabs}
\newcolumntype{.}{D{.}{.}{-1}}

\begin{document}

\newcommand{\titlename}{A Survey On Text-to-3D Contents Generation \\ In The Wild}
\title{\titlename}

\author{\IEEEauthorblockN{Chenhan Jiang}

\IEEEauthorblockA {
\textit{Department of Computer Science and Engineering} \\
\textit{The Hong Kong University of Science and Technology} \\
Hong Kong SAR \\
Email: \href{mailto:cjiangao@connect.ust.hk}{cjiangao@connect.ust.hk}
}
}

\IEEEtitleabstractindextext{

    \parbox{0.918\textwidth}{
\begin{abstract}
3D content creation plays a vital role in various applications, such as gaming, robotics simulation, and virtual reality. However, the process is labor-intensive and time-consuming, requiring skilled designers to invest considerable effort in creating a single 3D asset. To address this challenge, text-to-3D generation technologies have emerged as a promising solution for automating 3D creation. Leveraging the success of large vision language models, these techniques aim to generate 3D content based on textual descriptions. Despite recent advancements in this area, existing solutions still face significant limitations in terms of generation quality and efficiency.
In this survey, we conduct an in-depth investigation of the latest text-to-3D creation methods. We provide a comprehensive background on text-to-3D creation, including discussions on datasets employed in training and evaluation metrics used to assess the quality of generated 3D models. Then, we delve into the various 3D representations that serve as the foundation for the 3D generation process. Furthermore, we present a thorough comparison of the rapidly growing literature on generative pipelines, categorizing them into feedforward generators, optimization-based generation, and view reconstruction approaches. By examining the strengths and weaknesses of these methods, we aim to shed light on their respective capabilities and limitations. Lastly, we point out several promising avenues for future research.  With this survey, we hope to inspire researchers further to explore the potential of open-vocabulary text-conditioned 3D  content creation.

\end{abstract}
}
}

\maketitle
\IEEEdisplaynontitleabstractindextext
\IEEEpeerreviewmaketitle

\begin{figure*}[t]
\centering
\includegraphics[width=1\textwidth]{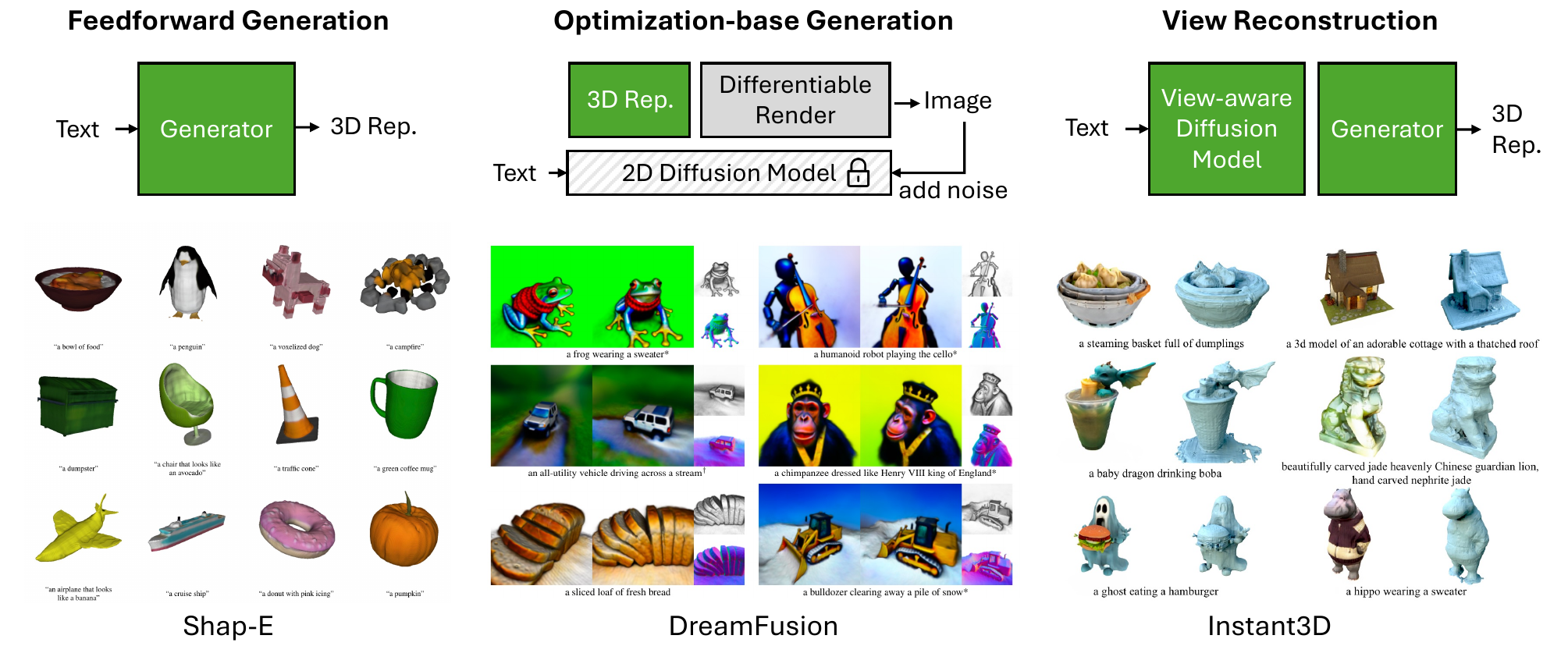}
\caption{In this survey, we investigate various text-to-3D content generation in the wild and categorize them delineated by algorithmic methodologies. Feedforward generation directly outputs 3D representations given text. Optimization-based generation optimizes parametric 3D representations using gradients from a 2D diffusion model. View reconstruction follows a text-to-images-to-3D paradigm. Representative 3D generation results are obtained from Shap-E~\cite{jun2023shape}, DreamFusion~\cite{dreamfusion22} and Instant3D~\cite{li2023instant3d}.}
\label{fig:intro}
\end{figure*}
\IEEEraisesectionheading{\section{Introduction}\label{sec:introduction}}

\IEEEPARstart{T}{he} demand for 3D content creation in industries such as gaming and filming has surged. However, manual creation of 3D assets requires specialized tools and expertise, posing a significant barrier to entry. To address this, there has been a growing interest in utilizing generative AI techniques for automated and high-quality 3D content generation. 
The use of natural language as a convenient tool for user interaction has emerged as a promising approach. As a result, the field of text-to-3D generation~\cite{nichol2022pointe,jun2023shape,dreamfusion22,magic3d22} has gained momentum, focusing on the development of technologies that utilize open-vocabulary text descriptions for automatic 3D content production. 
The diversity, quality and efficiency of text-to-3D generation methods have become crucial concerns for the community.

While recent advancements in large-scale Vision-Language Models (VLMs)~\cite{sanghi2022clip,rombach2022high,ramesh2022hierarchical,saharia2022photorealistic} have greatly enhanced open-vocabulary text-to-image generation, transitioning from 2D to 3D content generation presents unique challenges. Unlike 2D content, 3D content requires handling unstructured and unordered explicit 3D representations, which are not easily integrated into neural networks. Obtaining implicit 3D representations from sparse 3D data for generation is also challenging. Additionally, the available text-3D paired datasets~\cite{deitke2023objaverse,deitke2023objaversexl} consist of only 10 million examples, which pales in comparison to the vast amounts of internet-scale data used to train 2D counterparts, such as LAION-5B~\cite{schuhmann2022laion}. As a result, direct supervision on 3D data paired with natural language descriptions often fails to yield satisfactory results beyond the training distribution~\cite{autosdf2022,zhang20233dshape2vecset,cheng2023sdfusion}. Nevertheless, avoiding reliance on text-3D pairs presents optimization challenges~\cite{dreamfusion22,magic3d22,wang2023prolificdreamer} compounded by the complexity of the underlying 3D generative models, ultimately leading to substantial computational and training time requirements.
Moreover, Industrial design pipelines may not readily adopt current text-to-3D generation methods due to unreasonable topology and Janus problem in generated results. Overcoming these complexities necessitates innovative approaches and novel solutions to bridge the gap between 2D and 3D content generation.

While not as prominently featured as its 2D counterpart, 3D content generation has been steadily progressing with a series of notable achievements. The representative examples shown in~\cref{fig:intro} demonstrate the three main categories of current methods like Shap-E~\cite{jun2023shape}, DreamFusion~\cite{dreamfusion22} and Instant3D~\cite{li2023instant3d}. Unlike related surveys~\cite{shi2022deep,liao2024advances} addressing 3D generation with various conditions and 3D representation, we focus on a detailed categorization and discussion of text-conditioned 3D content generation with an open-vocabulary nature. 
In this survey, we first discuss the scope and related work of this survey in Section~\ref{sec:scope}. Section~\ref{sec:fundamentals} introduces the fundamentals of the 3D generation task, including formulations of various VLMs and common-used 3D datasets. Section~\ref{sec:3d_representation} introduces popular 3D representations. Then, we explore a wide variety of 3D generative pipelines, which can be divided into three categories based on their algorithmic methodologies: feedforward generation (Section~\ref{sec:feedforward_generation}), optimization-based generation (Section~\ref{sec:optimization_generation}) and view reconstruction (Section~\ref{sec:view_reconstruction}). \cref{fig:intro} and provide a summary of this categorization. The feedforward generation is similar to text-to-image methods, and we further break down the methods that are supervised with text-3D paired data and 2D VLMs. For optimization-based generation, we further explore the improvements by using enhanced 3D representation, improving optimization strategy, modifying SDS objective, and fine-tuning diffusion prior. For view reconstruction, we primarily focus on multi-view reconstruction that produces consistent and high-quality results. In the end, we outline open challenges, and conclude this survey. We hope this survey offers a systematic summary of 3D generation that could inspire subsequent work for interested readers.

\section{Scope of this survey}\label{sec:scope}

This survey examines methods for generating open-vocabulary text-based 3D content, with a focus on techniques that can produce diverse 3D representations aligned with the provided text descriptions, without restricting the vocabulary or categories of contents. The surveyed papers are primarily from major computer vision and computer graphics conferences/journals, and some preprints released on arXiv in 2024 which are discussed in future work.
In contrast to conventional text-to-3D shapes methods like~\cite{li2023diffusion,autosdf2022,autosdf2022,zhang20233dshape2vecset,cheng2023sdfusion}, which learn a cross-modal distribution by directly learning from limited text-3D pairs~\cite{chang2015shapenet}, these methods face challenges in acquiring 3D data and struggle to produce results beyond the training distribution. For readers interested in a review of such approaches, Shi et al.~\cite{shi2022deep} can be referred to.
Furthermore, this survey does not include image-to-3D generative methods without text-only conditions as presented in works, such as~\cite{liu2023zero,liu2024one}.

While the most relevant survey~\cite{li2023generative} briefly covers early techniques in text-to-3D generation, it does not include the latest methods and multiview reconstruction approaches. It should be emphasized that this survey provides an in-depth complement to existing surveys on 3D content generation\cite{liao2024advances,liu2024comprehensive}, as none of the previous surveys comprehensively cover text-conditioned 3D object generation and provides relevant insights.
\section{Fundamentals}\label{sec:fundamentals}

\subsection{Vison-Language Models}
Vision-Language Models (VLMs) have demonstrated impressive capabilities in generating remarkable outputs based on text prompts in the 2D field, inspiring the development of similar techniques or serving as guidance models in the 3D domain.

\noindent\textbf{CLIP. }Contrastive Language-Image Pre-training (CLIP) is a notable example that exhibits strong cross-modal matching between images and language. It is commonly used to determine whether an image and its associated caption are a suitable match or not. 
\begin{figure}[b]
\centering
\includegraphics[width=0.95\linewidth]{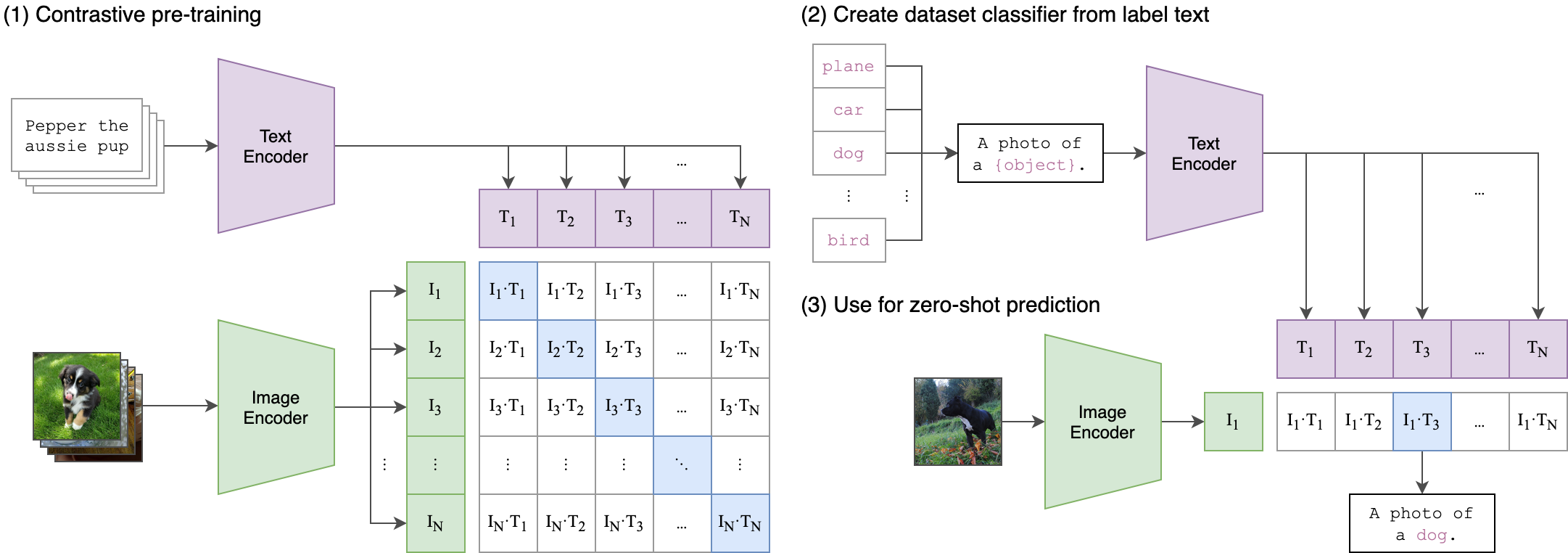}
\caption{The structure of CLIP~\cite{radford2021learning}.}
\label{fig-foundclip}
\end{figure}
CLIP consists of a text encoder and image encoder which project text and images onto an aligned latent space. The model is trained by minimizing the contrastive loss on 400 million text-image pairs. And many methods calculate the similarity between a pair of image and text embedding from corresponding CLIP encoders. \cref{fig-foundclip} illustrates the structure of CLIP.

\begin{table*}[t]
\centering
\tiny   
\resizebox{0.95\textwidth}{!}{
\begin{tabular}{c|ccccc}
\toprule
Dataset & Objects & Categories & Text & Data Source & Text Source\tabularnewline\midrule 
ShapeNet~\cite{chang2015shapenet} & 51K & 55 & - & - & - \tabularnewline
Objaverse~\cite{deitke2023objaverse} & 800k & open & - & - & - \tabularnewline
Objaverse-XL~\cite{deitke2023objaversexl} & 10.2M & open & - & - & - \tabularnewline
\bottomrule
Text2Shape~\cite{chen2019text2shape} & 15K & 2 & 75K & ShapeNet & Human\tabularnewline
ShapeGlot~\cite{shapeglot} & 5K & 1 & 79K & ShapeNet & Human \tabularnewline
ShapeTalk~\cite{achlioptas2023shapetalk} & 36K & 30 & 536K & ShapeNet & Human\tabularnewline
OpenShape~\cite{liu2024openshape} &  876K & open & 876K & Objaverse & VLM\tabularnewline
Cap3D~\cite{luo2023scalable} & 1M & open & 1M & Objaverse, Objaverse-XL & VLM\tabularnewline
Shap-E~\cite{jun2023shape}& \textgreater 1M & open & 120K & Point-E & Human\tabularnewline
\bottomrule
\end{tabular}
}

\caption{Commonly used 3D datasets in text-to-3D generation. The '-' symbol denotes the absence of an involved attribute. Note that Shap-E is the largest text-3D paired dataset, although it is not open-source. Furthermore, Cap3D is an available dataset that provides multiple versions of 3D object-caption paired data, with varying quality based on filtering. }
\label{tab-dataset}
\end{table*}
\begin{figure}[t]
\centering
\includegraphics[width=0.95\linewidth]{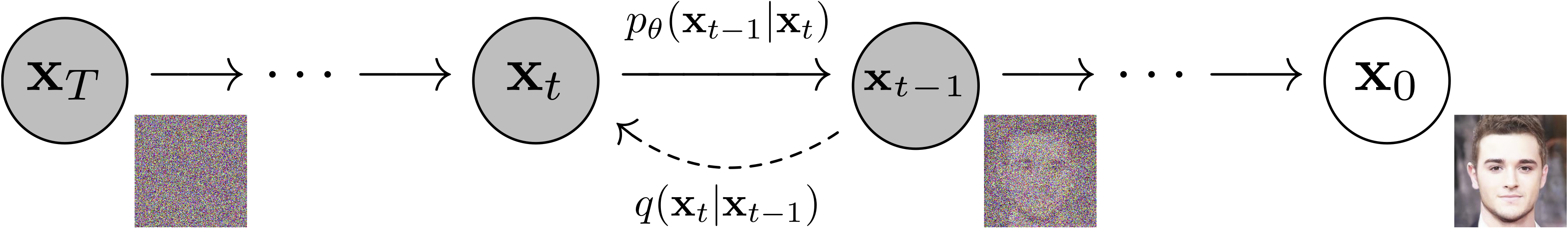}
\caption{The process of DDPM~\cite{ho2020denoising}.}
\label{fig-founddm}
\end{figure}

\noindent\textbf{Diffusion Models. }Recently, Diffusion Models (DMs)~\cite{rombach2022high} have garnered significant attention and achieved impressive results in text-conditional generation within the 2D domain. Also referred to as denoising diffusion probabilistic models (DDPMs) shown in~\cref{fig-founddm}, DMs comprise a forward process $q$ that gradually introduces noise $\epsilon$ to the input data $x_0$ according to a noise schedule $\beta_{1:T}$ spanning $T$ time steps.
\begin{equation}\small
\begin{split}
    & q(\mathbf{x}_t | \mathbf{x}_{t-1})=\mathcal{N}(\mathbf{x}_t;\sqrt{1-\beta}\mathbf{x}_{t-1},\beta_t\mathbf{I}),\\
    & q(\mathbf{x}_{1:T}|\mathbf{x}_0)=\prod_{t=1}^Tq(\mathbf{x}_t|\mathbf{x}_{t-1})
\end{split}
\end{equation}
They also encompass a reverse process or generative model $p_\Phi$ that iteratively denoises the Gaussian distribution $x_T$ to generate an image from the desired data distribution. The generative model $p_\Phi$ is trained using a (weighted) Evidence Lower Bound (ELBO), which can be simplified to a weighted denoising score matching objective for the parameters $\Phi$~\cite{kingma2021variational}.
\begin{equation}
L_{DM}=\mathrm{E}_{t\sim\mathcal{U}(1,T), \epsilon\sim\mathcal{N}(\textbf{0}, \textbf{I})}[w(t)||\epsilon_{\Phi}(\mathbf{x}_t,t)-\epsilon)||^2],
\label{eq-founddm}
\end{equation}
where $w(t)$ is a weighting term and $\epsilon_\Phi$ is a network used to predict noise, typically implemented as a UNet~\cite{ronneberger2015u}. For text-to-image diffusion models, they learn $\hat\epsilon_{\Phi}$ conditioned on text embeddings $y$ with classifier-free guidance (CFG) scale $\omega$~\cite{ho2022classifier}:
\begin{equation}
\hat\epsilon_{\Phi}=(1+\omega)\epsilon_{\Phi}(\mathbf{x}_t,t,y)-\omega\epsilon_{\Phi}(\mathbf{x}_t,t,\emptyset)
\end{equation}

\subsection{Datasets}

Nowadays, data plays a crucial role in ensuring the success of algorithms. A well-curated dataset can significantly enhance a model's robustness and performance. On the contrary, noisy and flawed data may cause model bias that requires considerable effort in algorithm design to rectify. In this part, we introduce the commonly used datasets and their derivative works for text-to-3D generative methods. The summary comparison can be found in~\cref{tab-dataset}.

\noindent\textbf{ShapeNet. }
ShapeNet~\cite{chang2015shapenet} is introduced to build a large-scale repository of 3D CAD models of objects. The core of ShapeNet covers 55 common object categories with about 51,300 models that are manually verified category and alignment annotations. However, one limitation of this dataset is the absence of text descriptions for each object. To address this gap, subsequent works~\cite{shapeglot,chen2019text2shape,achlioptas2023shapetalk}  have supplemented ShapeNet with human-annotated captions. For example, Text2Shape~\cite{chen2019text2shape} provides paired text and 3D object data specifically for the table and chair categories. ShapeGlot~\cite{shapeglot} and ShapeTalk\cite{achlioptas2023shapetalk} release discriminative text where one object is selected multiple objects. Another method~\cite {xue2023ulip,xue2023ulip2} explores automatically adding captions to the rendered images of ShapeNet using an image caption model~\cite{li2023blip}. Despite these efforts to enhance the dataset, ShapeNet still suffers from limitations in scalability and diversity, restricting its use for training arbitrary text-to-3D generative models.

\noindent\textbf{Objaverse series. }
Objaverse~\cite{deitke2023objaverse} introduces a substantial corpus of 3D objects, comprising over 800,000 3D assets collected from various 3D model repositories. To expand the dataset further, Objaverse-XL~\cite{deitke2023objaversexl} further extends Objaverse to a larger 3D dataset of 10.2M unique objects from a diverse set of sources. These large-scale 3D datasets have the potential to facilitate large-scale training and boost the performance of 3D generation. These extensive 3D datasets hold immense potential for facilitating large-scale training and enhancing the performance of 3D generation models. However, it is worth noting that the text descriptions within these datasets contain considerable noise, and undesired shapes may be present. To address this issue, OpenShape~\cite{liu2024openshape} and Cap3D~\cite{luo2023scalable} provide filtered versions of the dataset. These filtered versions utilize large-language models to generate more informative text descriptions for shapes. These filtered datasets have been widely used in text-to-3D generative methods~\cite{yi2023gaussiandreamer,li2023sweetdreamer,qiu2023richdreamer}.
\section{3D Representations}\label{sec:3d_representation}

\begin{figure*}[t]
\centering
\includegraphics[width=1\linewidth]{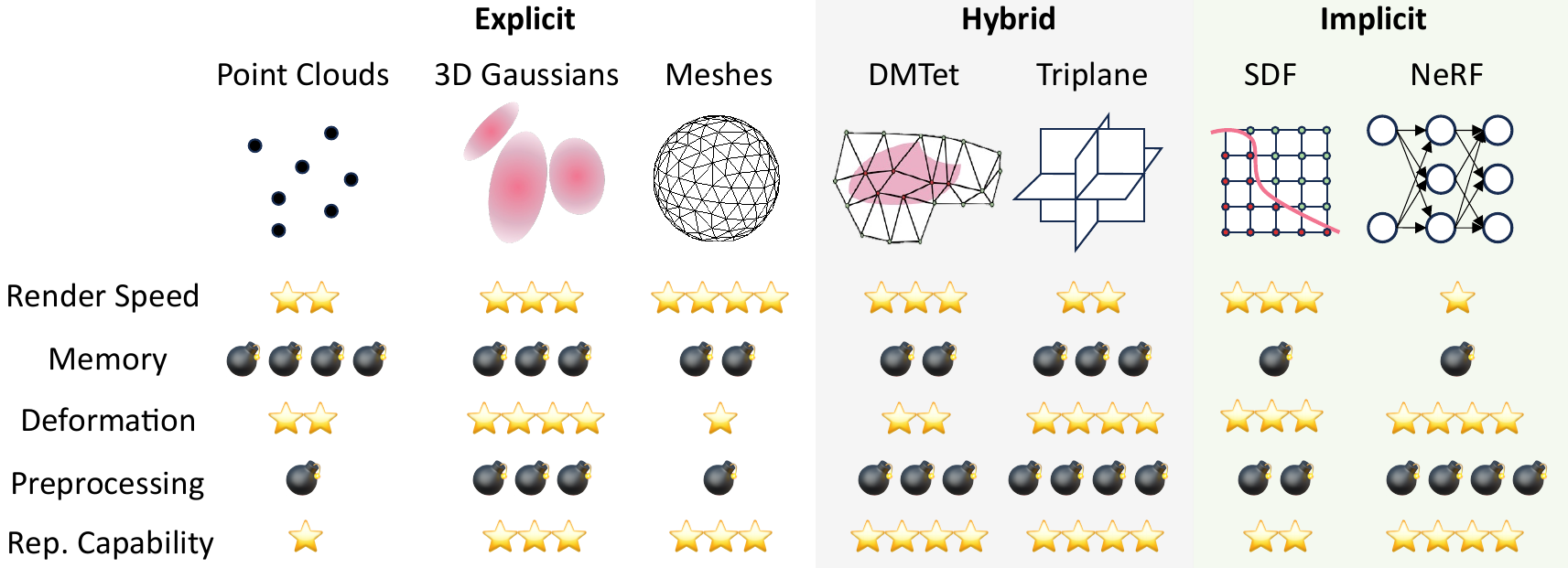}
\caption{Comparison of different representations with regard to rendering speed, memory usage with increasing resolution, shape deformation, the time of data preprocessing and representation capacity for arbitrary geometry. A larger number of $\bigstar$ and a smaller number of \bomb\ indicate better performance.}
\label{fig-3drep}
\end{figure*}
In the realm of text-to-3D content generation, selecting an appropriate representation for 3D content generation is crucial. The generation process usually entails utilizing a 3D representation along with a differentiable rendering algorithm to construct 3D content and generate corresponding 2D images. Different methods employ different approaches to supervise the generation process. Some methods directly supervise the 3D content within the 3D representation, whereas others supervise the rendered images resulting from the 3D representation. In the subsequent sections, we categorize the 3D representations into three main groups: explicit, implicit, and hybrid. \cref{fig-3drep} shows the comparison among different 3D representation.

\subsection{Explicit 3D Representation}
Explicit 3D representations, such as point-based or polygon-based structures, are characterized by a fixed number of 3D elements. This property allows for the preservation of geometry and facilitates seamless integration with differentiable rasterization techniques. However, the fixed nature of these representations poses challenges in terms of scalability and flexibility compared to implicit representations.

\noindent\textbf{Point Clouds. }
A point cloud is an unstructured collection of elements in Euclidean space that represents discrete points in a three-dimensional environment. These points can have additional attributes such as colors and normals, and in some cases, depth and normal maps can be considered as specific instances of point cloud representations. Point clouds are commonly obtained directly from depth sensors, making them widely used in various tasks related to 3D scene understanding. However, despite their easy acquisition, point clouds present challenges for traditional 2D neural networks due to their irregularity, making it difficult to process them effectively. Moreover, the disconnected and unstructured nature of point clouds introduces ambiguity in terms of the underlying geometry.
Considering these limitations, only the Point-E approach~\cite{nichol2022pointe} adopts point clouds as the chosen representation and generates coarse shapes by using a limited number of points. This design choice is influenced by the fact that constructing high-resolution shapes requires a significant number of points, resulting in a substantial consumption of GPU memory.

\noindent\textbf{3D Gaussians}
3D Gaussians have emerged as an improved alternative to point clouds and have gained widespread use in 3D reconstruction tasks~\cite{kerbl20233d}. Their efficient rendering capabilities and flexibility make them highly appealing. Instead of using point clouds, 3D Gaussians represent objects as a collection of anisotropic Gaussian distributions parameterized by their positions, covariances, colors, and opacities. During the rendering process, these 3D Gaussians are projected onto the imaging plane of the camera, and the resulting 2D Gaussians are assigned to individual tiles.
The application of 3D Gaussians has also extended to the text-to-3D domain~\cite{yi2023gaussiandreamer,tang2023dreamgaussian,EnVision2023luciddreamer,zuo2024videomv,tang2024lgm,chen2023text,chen2024v3d}.While 3D Gaussians offer fast convergence, they are sensitive to initialization and can exhibit unstable optimization. Consequently, in optimization-based generative pipelines, such as those observed in ~\cite{tang2023dreamgaussian, yi2023gaussiandreamer, chen2023text}, the issue of server multi-janus problem has been identified.
Some methods overcome these issues by applying reconstruction-based 3D generative pipline~\cite{zuo2024videomv,tang2024lgm,chen2024v3d}, or considering providing a more favorable initial model~\cite{EnVision2023luciddreamer}.

\noindent\textbf{Mesh. }
Meshes are a popular 3D representation used in computer vision and graphics. They consist of vertices, edges, and faces, and are efficient in terms of memory usage and scalability. Meshes only encode the surfaces of objects, making them more memory-efficient than voxels. Compared to point clouds, meshes provide explicit connectivity information and are suitable for geometric transformations. They can also encode textures conveniently. Recently, differentiable mesh rendering methods~\cite{Laine2020diffrast,hasselgren2021appearance,munkberg2022extracting} have been developed to update properties defined on meshes and enable meshes to be rasterized for gradient-based optimization.
However, working with and generating 3D meshes can present challenges due to the irregular data structure and complexity of predicting vertex positions and topology. As of now, there are no text-to-3D generation methods that directly optimize mesh representations. Existing solutions typically involve transforming intermediate representations into meshes through surface reconstruction techniques (for point-like representations) or iso-surface extraction techniques~\cite{lorensen1998marching, 2022dmtet} (for implicit representations).

\subsection{Implicit 3D Representation}
Implicit representations have gained popularity as the preferred 3D representation for view synthesis, 3D reconstruction, and a variety of other applications in computer graphics and computer vision. These representations involve constructing a mapping function that describes the properties of a 3D space, either through mathematical formulations or neural networks.
In contrast to explicit scene representations that focus on object surfaces, implicit representations have the ability to define the entire volume of a 3D object. They offer the flexibility to represent 3D scenes or objects at arbitrary resolutions and with improved memory efficiency. 
Rendering an image from an implicit representation typically involves volume rendering~\cite{kajiya1984ray}, which employs ray casting and samples multiple points along each ray. However, sampling a set of points along all rays can lead to slow rendering speeds.
While implicit representations excel in shape modeling, their lack of ground-truth data in implicit representation format hinders their direct use in supervised generative pipelines. Consequently, most approaches utilizing implicit representations rely on optimization-based pipelines~\cite{dreamfusion22, wang2023prolificdreamer, zhu2023hifa, chen2024vp3d}.

\noindent\textbf{Signed Distance Field. }
The Signed Distance Function (SDF), defines a 3D surface as the zero-level set of a distance field, where each point in space is assigned a value corresponding to its signed shortest distance to the surface. Specifically, given SDF function $f$ maps a coordinate $\mathbf{x}$ to a scalar $f(\mathbf{x})=d$, $|d|$ is the distance of $\mathbf{x}$ to the nearest point on the surface of the shape, and $d<0$ means point $\mathbf{x}$ is outside of the shape. According to this definition, the level set $f(\mathbf{x})=0$ defines the surface of the shape. Therefore, SDF allows for efficient operations like constructed solid geometry by utilizing iso-surface extracting techniques like marching cubes~\cite{lorensen1998marching} or marching tetrahedra~\cite{doi1991efficient}.

\noindent\textbf{Neural Radiance Field. }
The Neural Radiance Field (NeRF)~\cite{mildenhall2021nerf} represents a 3D content as a continuous volumetric function $f$, which utimizes MLPs to map the position $\mathbf{x}\in\mathbb{R}^3$ and the viewpoint $\mathbf{d}\in\mathbb{R}^2$ to a density $\sigma$ and color $c$: $f(\mathbf{x}, \mathbf{d})=(\sigma,c)$. To render a pixel in images, NeRF casts a single ray $\mathbf{r}(t)=\mathbf{o}+t\mathbf{d}$ and samples points $t_i$ to accumulate into the pixel color $C(\mathbf{r})$ via volume rendering:
\begin{equation}
C(\mathbf{r})=\sum_i T_i\alpha_i c_i,\text{where }T_i=\text{exp}(=\sum_{k=0}^{i-1}\sigma_k(t_i-t_{i-1})) 
\end{equation}
where $\alpha_i$ indicates the opacity of the sampled point and $T_i$ quantifies the probability of the ray traveling from $t_0$ to $t_i$ without encountering other particles. NeRF is the most popular representation used in optimization-based generative methods~\cite{dreamfusion22, wang2023prolificdreamer, zhu2023hifa, chen2024vp3d} due to its flexibility. However, the biggest challenge is slow rendering speeds.

\subsection{Hybrid Representation}
Given the respective advantages and disadvantages of each representation, hybrid representations have been proposed as a means to complement and combine their strengths. Many of these hybrid representations primarily concentrate on the fusion of explicit and implicit representations. Explicit representations provide explicit control over the geometry. On the other hand, they are restricted by the resolution and topology. Implicit representations allow for the modeling of complex geometry.

\noindent\textbf{DMTet. }
DMTet~\cite{2022dmtet} is a hybrid three-dimensional surface representation that combines explicit tetrahedral grids and implicit SDF to create a versatile and efficient model. The 3D space is divided into dense deformable tetrahedral grids $\{{V_t}\}^T$. For each vertex $v_i\in V_t$, a network $f$ is used to predict SDF value $s(v_i)$ and a position offset $\Delta v_i$ by: $(s(v_i),\Delta v_i)=f(v_i)$. And it can be easily transform to meshes through differentiable March Tetrahedral layer during training, yielding a fast and high-resolution rendering.

\noindent\textbf{Triplane. }
Another promising hybrid approach in text-to-3D content generation is the triplane representation, introduced by EG3D~\cite{chan2022efficient}, where the 3D information is stored in three axis-aligned orthogonal 2D feature planes $\textbf{f}_{xy},\textbf{f}_{xz},\textbf{f}_{yz}\in\mathbb{R}^{N\times N\times C}$ with a spatial resolution $N\times N$ and feature channels $C$. To predict the color $c$ and density values $\sigma$ at each point $\mathbf{x}$, a MLP decoder takes in the aggregated 3D features from the three planes.
\begin{equation}
(\sigma, c) = \text{MLP}(\textbf{f}_{xy}(\mathbf{x})+\textbf{f}_{xz}(\mathbf{x})+\textbf{f}_{yz}(\mathbf{x}))
\end{equation}
This representation consumes less memory than voxel-based NeRF, and they allow fast rendering at the same time.
\section{Feedforward Generation}\label{sec:feedforward_generation}

Feedforward text-to-3D generation involves training a generator to capture the alignment between text and 3D representations. This approach draws inspiration from various techniques such as generative adversarial networks~\cite{creswell2018generative,gao2022get3d}, autoregressive networks~\cite{mittal2022autosdf}, and diffusion models~\cite{rombach2022high}, enabling the generator to produce 3D representations directly from text descriptions within minutes.
However, previous feedforward 3D models~\cite{mittal2022autosdf,cheng2023sdfusion,sanghi2022clip} have faced limitations in handling a wide range of categories due to the scarcity of available text-3D pairs. To address this challenge and enable open-vocabulary 3D generation, two approaches have emerged: leveraging large-scale 3D datasets and utilizing well-trained 2D VLMs.

\begin{table*}[t]
\centering
\resizebox{1\textwidth}{!}{
\begin{tabular}{c|ccccccc}
\toprule
Method & Latent Emb. & 3D Rep. & Input & Encoder & Decoder & Loss & Data \tabularnewline\midrule 
3DGen~\cite{gupta20233dgen} & Triplane & DMTet & PCL & PointNet-UNet & UNet-MLP & render & SN, OBJ \tabularnewline
Shap-E~\cite{jun2023shape} & NeRF Param. & NeRF & PCL + Image& Transformer & - & render & Shap-E \tabularnewline
Michelangelo~\cite{zhao2024michelangelo} & Shape Emb. & Occupancy Field & PCL & Transformer & Transformer & contrastive+BCE & SN, OBJ \tabularnewline
\bottomrule
\end{tabular}
}

\caption{Comparison of the VAE training in feedforward generation for learning from 3D data. This involves training a 3D encoder and decoder to project 3D data, such as point clouds, into a latent space. In the case of 3DGen~\cite{gupta20233dgen} and Shap-E~\cite{jun2023shape}, a rendering-based reconstruction loss (render) is utilized. On the other hand, Michelangelo~\cite{zhao2024michelangelo}  employs a combination of text-image-shape contrastive loss and binary cross-entropy loss (BCE). PCL, SN, OBJ are the abbreviations of Point Clouds, ShapeNet, Objaverse respectively.}
\label{tab-feedforward1}
\end{table*}
\subsection{Learning From 3D Datasets}~\label{sec-fffrom3d}
\begin{figure*}[t]
\centering
\includegraphics[width=1\linewidth]{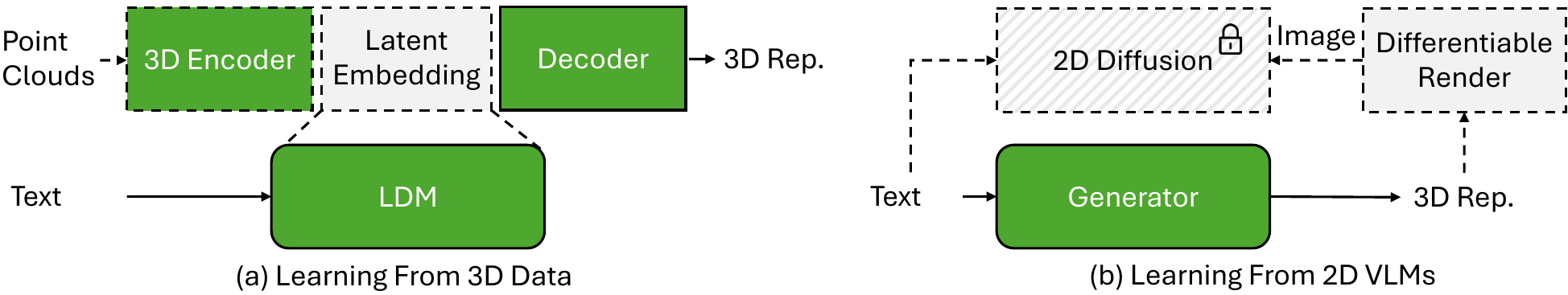}
\caption{The pipelines of feedforward 3D generation models. The dotted line parts are only used during training.}
\label{fig-ff}
\end{figure*}

The release of large-scale 3D datasets~\cite{deitke2023objaverse,deitke2023objaversexl} has significantly influenced the development of feedforward text-to-3D generation. However, obtaining supervision from 3D data presents challenges. Explicit 3D representation requires extensive memory resources to handle high-resolution 3D data, while implicit 3D representation cannot be directly obtained from the data itself. To overcome these obstacles, many methods adopt a two-stage pipeline shown in~\cref{fig-ff} (a). First, a VAE model~\cite{kingma2021variational} is trained, consisting of a 3D encoder and decoder to produce a latent embedding. Then, a latent diffusion model (LDM) as the generator is trained based on latent embedding. This survey focuses on the design and training of the VAE model, and a summary of the VAE model comparison is presented in~\cref{tab-feedforward1}.

\begin{figure}[t]
\centering
\includegraphics[width=1\linewidth]{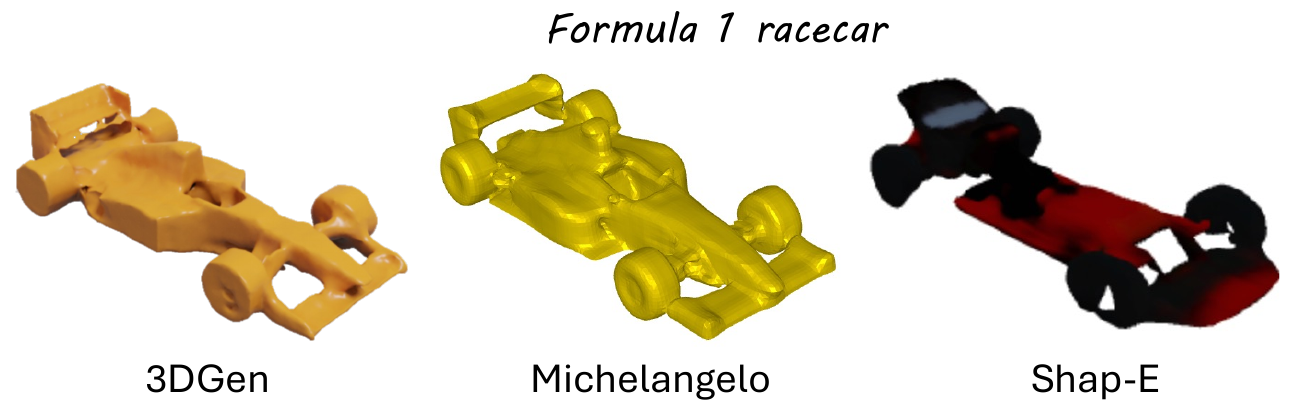}
\caption{Qualitative results of feedforward generation from 3D dataset. Compared with DMTet and occupancy representation, NeRF used in Shap-E~\cite{jun2023shape} tends to produce holes.}
\label{fig-ffdemo1}
\end{figure}

3DGen~\cite{gupta20233dgen}, utilizes PointNet to extract point cloud features and employs UNet to generate a triplane as the latent embedding. The decoder refines the triplane latent using UNet and predicts attributes of a DMTet representation with a MLP. VAE training is conducted using a rendering-based reconstruction loss with differentiable rasterization and the marching tetrahedra algorithm. Another method, Shap-E~\cite{jun2023shape}, also employs a rendering-based reconstruction loss for training the VAE model. However, they use NeRF as the representation and employ a transformer-based 3D encoder to predict its parameters, acting as weight matrices for an MLP. It should be noted that directly learning a conditional generator, such as 3DGen and Shap-E, from the conditions requires a large amount of data and may result in low-quality and less diverse results due to the significant distribution gap between the 3D space and the image/text space.
Michelangelo~\cite{zhao2024michelangelo} addresses the distribution gap by training the VAE model to align language, image, and 3D shape in the latent space using a contrastive loss and a frozen CLIP model~\cite{radford2021learning}, similar to other 3D representation learning methods~\cite{xue2023ulip,xue2023ulip2,zeng2023clip2}. In addition to the contrastive loss, a binary cross-entropy loss is used to supervise the occupancy field produced by the decoder. 
\cref{fig-ffdemo1} provides a qualitative comparison of these methods, where Michelangelo produces structures that closely match the conditions due to the alignment among text, image, and shape. Shap-E benefits from training on an expanded text-3D paired dataset, although this dataset is not publicly available. However, it is worth noting that NeRF used in Shap-E tends to produce holes.

While feedforward generative methods learning from 3D data can produce geometry-accurate results, they often lack high-frequency structure and detailed texture, as demonstrated in~\cref{fig-ffdemo1}. Nonetheless, these methods can serve as valuable initializations for follow-up methods~\cite{liu2023sherpa3d,yi2023gaussiandreamer}.

\subsection{Learning From 2D VLMs}~\label{sec-fffromvlm}
Despite the recent introduction of larger 3D datasets, they remain significantly smaller compared to contemporary image-text datasets, which typically contain billions of examples or have limited diversity and texture, as seen in studies~\cite{jun2023shape,gupta20233dgen,zhao2024michelangelo}. An alternative approach is to leverage well-trained 2D VLMs, such as diffusion models~\cite{rombach2022high} and CLIP~\cite{radford2021learning}, to capture the 3D shape distribution. \cref{fig-ff} (b) illustrates the general pipeline for this approach. However, it is important to note that this method necessitates the careful construction of prompt sets to adapt 2D VLMs to the task of 3D generation.

ATT3D~\cite{lorraine2023att3d} is the first method to train a generator using amortized optimization~\cite{amos2023tutorial} across multiple prompts. The generator in ATT3D comprises a mapping network that takes a text prompt as input and an Instant-NGP~\cite{mueller2022instant} model to produce a NeRF representation. The NeRF representation is then used to render multi-view images, and the mapping network is trained using the SDS loss computed via a 2D diffusion model. During training, text embeddings are interpolated to amortize over the text, facilitating smooth interpolations between different text prompts. Prompt sets in ATT3D are built using the template: "a $\{\tt{animal}\}$ $\{\tt{activity}\}$ $\{\tt{theme}\}$ $\{\tt{hat}\}$", where activities, themes, and hats can be combined in various ways. While ATT3D shows promising results on trained prompts, it struggles when faced with general prompts. The simple architecture of ATT3D has limited capacity and lacks strong inductive biases for 3D generation, making it challenging to scale with dataset size and rendering resolution. Consequently, the method is limited to small-scale prompt sets (100s-1000s) and low-fidelity textures. Subsequent works~\cite{li2023instant3dtext,qian2024atom,chen2023et3d} address these limitations by expanding the network architecture and replacing the NeRF representation with Triplane, resulting in improved quality. ET3D specifically trains a GAN model from a view-aware diffusion model~\cite{shi2023mvdream} to capture the real data distribution, mitigating issues such as over-saturation, over-smoothing, low diversity, and the multi-face Janus problem. HyperFields~\cite{babu2023hyperfields} trains a hypernet to record the NeRF parameters from individual training. Nevertheless, all of these methods still face challenges associated with the multi-Janus problem and the limited prompt sets. In contrast, Latte3D~\cite{xie2024latte3d} leverages 3D data and a 3D-aware diffusion model~\cite{shi2023mvdream} to ensure geometry consistency. Notably, Latte3D demonstrates scalability to an impressive order of 100,000 prompts by incorporating rule-based text generation and ChatGPT~\cite{openai2023gpt}, using the template "a $\{\tt{object }\}$ in $\{\tt{style}\}$ is $\{\tt{doing}\}$".
\section{Optimization-based Generation}\label{sec:optimization_generation}

\begin{figure}[b]
\centering
\includegraphics[width=0.95\linewidth]{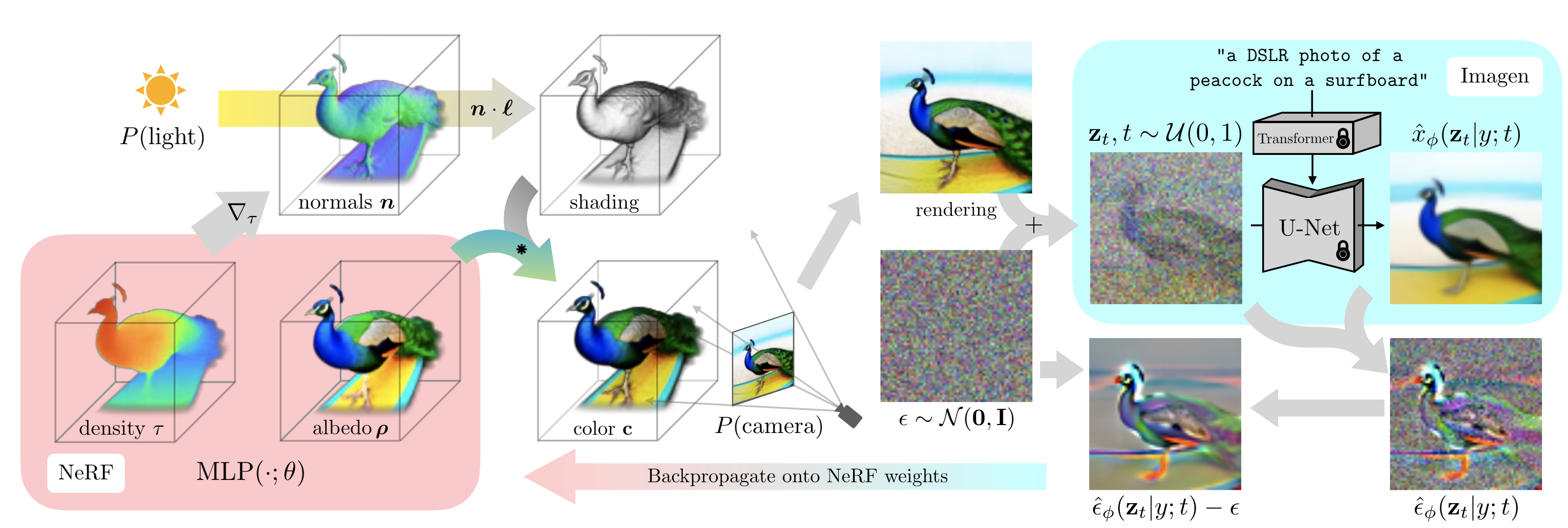}
\caption{DreamFusion~\cite{dreamfusion22} proposes to optimize NeRF with score distillation sampling loss from a 2D diffusion prior.}
\label{fig-dreamfusion}
\end{figure}

In the domain of text-to-3D content generation, researchers have increasingly turned to optimization-based approaches inspired by the success of text-to-image diffusion models~\cite{rombach2022high}. These methods offer a viable alternative by sidestepping the need for large-scale text-3D datasets when training scalable 3D generators. A pioneering work~\cite{dreamfusion22} introduces a crucial technique called Score Distillation Sampling (SDS), shown in~\cref{fig-dreamfusion}. SDS utilizes diffusion priors as score functions to guide the runtime optimization of a NeRF representation. Concurrently, Wang et al.~\cite{wang2023score} propose a similar technique that leverages the chain rule and the learned gradients of a diffusion model which backpropagates the scores from the diffusion model through the Jacobian of a differentiable renderer. More specifically, given a 3D representation with learnable parameters $\theta$ and a pre-trained 2D diffusion model with noise prediction network $\epsilon_{\Phi}(\text{x}_t, t, y)$, SDS optimizes $\theta$ by minimizing the KL-divergence as follows: 
\begin{equation}\small
    \min\limits_{\theta}D_{KL}(q_t^{\theta}(\mathbf{x}_t|c, y) || p_t(\mathbf{x}_t|y)).
\label{eq-KLSDS}
\end{equation}
Here, $p_t(\text{x}_t|y)$ is the image distribution sampled from diffusion model, $q_t^{\theta}(\mathbf{x}_t|c, y)$ is the distribution of rendered image $\text{x}_t=g(\theta, c)$ with respect to camera pose $c$ at timestep $t$ of the forward diffusion process, where $g$ is the renderer. 
To solve~\cref{eq-KLSDS}, the score distillation function is derived as:
\begin{equation}\small
\begin{split}
    \nabla_{\theta}L_{SDS}(\theta) & \triangleq \mathrm{E}_{t,\mathbf{}{x}}[w(t)\frac{\sigma_t}{\alpha_t}\nabla_{\theta}KL(q_t^{\theta}(\mathbf{x}_t|c, y)||p_t(\mathbf{x}_t|y))]\\
    & \triangleq \mathrm{E}_{t,\epsilon_{\Phi}}[w(t)(\hat{\epsilon}_{\Phi}(\mathbf{x}_t, t, y)-\epsilon)\frac{\delta g(\theta, c)}{\delta \theta}],
\end{split}
\label{eq-SDS}
\end{equation}
where $w(t)$ denotes the time-dependent weighting function, and the classifier-free guidance (CFG)~\cite{ho2022classifier} trick is employed on noise with scale $s$:
\begin{equation}
    \hat{\epsilon}_{\Phi}:= (1+s)\epsilon_{\Phi}(\mathbf{x}_t, t, y) - s\epsilon_{\Phi}(\mathbf{x}_t, t, \emptyset)
\label{eq-cfg}
\end{equation}

\begin{table*}[t]
\resizebox{1\textwidth}{!}{
\centering
\begin{tabular}{c|ccccccc}
\toprule
Method & 3D Rep. & Objective & Motivation & Solution Type & 3D Prior & Resolution & Infer Time (min.)\tabularnewline\midrule 
3DFuse~\cite{seo2023let} & NeRF & SDS/SJC & Janus & Opt. Strategy & Point-E & 64 & -\tabularnewline
Prep-neg~\cite{armandpour2023re} & NeRF & SDS & Janus & Opt. Strategy & - & - & - \tabularnewline
Sherpa3D~\cite{liu2023sherpa3d} & DMTet & SDS & Janus & Opt. Strategy & Shap-E & 512$^2$ & 25 RTX3090 \tabularnewline
Consistent3D~\cite{wu2024consistent3d} & NeRF/DMTet/\textbf{Gaussians} & CDS & Janus & Improve SDS & - & 512$^2$ & 15 A100\tabularnewline
MVDream~\cite{shi2023mvdream} & NeRF & SDS & Janus & View-aware DM & - & 256$^2$ & 40 A100 \tabularnewline
PI3D~\cite{liu2023pi3d} & Triplane & SDS & Janus & View-aware DM & - & 256$^2$ & 3 A100\tabularnewline
RichDreamer~\cite{qiu2023richdreamer} & \textbf{NeRF}/DMTet & SDS & Janus & View-aware DM & - & 512$^2$ & 90 A100\tabularnewline
SweetDreamer~\cite{li2023sweetdreamer} & \textbf{NeRF}/DMTet & SDS & Janus & View-aware DM & - & 64$^2$ & 60 on 4 A100\tabularnewline
EfficientDreamer~\cite{zhao2023efficientdreamer} & NeuS-DMTet & SDS-VSD & Janus & View-aware DM & - & 64$^2$-512$^2$ & -\tabularnewline\bottomrule
DreamTime~\cite{huang2023dreamtime} & NeRF & SDS & Quality & Opt. Strategy & - & 64$^2$ & - \tabularnewline
Magic3D~\cite{huang2023dreamtime} & NeRF-DMTet & SDS & Quality & Opt. Strategy & - & 64$^2$-512$^2$ & 40 on 8 A100 \tabularnewline
Fantasia3D~\cite{chen2023fantasia3d} & DMTet & SDS & Quality & Opt. Strategy & - & 512$^2$ & 30 on 8 RTX3090 \tabularnewline
MATLABER~\cite{xu2023matlaber}& DMTet & SDS & Quality & Opt. Strategy & - & 512$^2$ & - \tabularnewline
HiFA~\cite{zhu2023hifa}& NeRF & SDS & Quality & Opt. Strategy & - & 512$^2$ & - \tabularnewline
Yu et al.~\cite{yu2023text}& NeRF-DMTet & CSD & Quality & Improve SDS & - & 64$^2$-512$^2$ & -\tabularnewline
LucidDreamer~\cite{EnVision2023luciddreamer}& Gaussians & ISM & Quality & Improve SDS & Point-E & 512$^2$ & 35 on 1 A100 \tabularnewline
ProlificDreamer~\cite{zhu2023hifa}& NeRF & VSD & Quality & Improve SDS & - & 64$^2$-512$^2$ & 560 on 8 A100 \tabularnewline\bottomrule
DreamPropeller~\cite{zhou2023dreampropeller}& \textbf{NeRF}/DMTet & SDS/\textbf{VSD} & Acceleration & Opt. Strategy & - & 64$^2$-512$^2$ & 130 on 8 A100 \tabularnewline
DreamGuassian~\cite{tang2023dreamgaussian} & Gaussians & SDS & Acceleration & Enhanced 3D Rep. & - & 128$^2$-1024$^2$ & 2 on 1 V100 \tabularnewline
GaussianDreamer~\cite{yi2023gaussiandreamer} & Gaussians & SDS & Acceleration & Enhanced 3D Rep. & Shap-E & 1024$^2$ & 15 on 1 RTX 3090 \tabularnewline
GSGEN~\cite{chen2023text} & Gaussians & SDS & Acceleration & Enhanced 3D Rep. & Point-E & 512$^2$ & 40 on 1 RTX 3090 \tabularnewline

\bottomrule
\end{tabular}
}

\caption{Summary of optimization-based generative methods. The "Objective" column denotes the optimization objective employed. The "Motivation" column indicates the primary challenge each method focuses on. The "Resolution" column represents the resolution of the rendered image from the 3D representation. Certain methods apply to various 3D representations or optimization objectives, which are separated by "/". The training time is reported for the method in bold. Additionally, some methods adopt a coarse-to-fine strategy with different objectives and rendering resolutions, which are connected by "-". Inference time of producing one object is taken from the respective papers, except for ProlificDreamer~\cite{wang2023prolificdreamer}, which was from DreamPropeller~\cite{zhou2023dreampropeller}.}
\label{tab-opt}
\end{table*}

Despite its popularity, empirical observations have shown that SDS often encounters issues such as multi-Janus problem, over-smoothing and time-consuming, which significantly hampers the practical application of high-fidelity 3D generation. Following works continue to solve these problems and can be divided into four aspects: Using enhanced 3D representation (\cref{sec-opt3dpre}), improving optimization strategy (\cref{sec-optop}), modifying SDS objective (\cref{sec-optsds}), and fine-tuning diffusion prior (\cref{sec-optft}). The whole summary and comparison can be found in~\cref{tab-opt}.

\subsection{Enhanced 3D Representation}\label{sec-opt3dpre}
Various approaches have been proposed to improve the render speed and fidelity of optimization-based text-to-3D generation. In this part, we discuss the utilization of DMTet and 3D Gaussians as enhanced representations, addressing the limitations of original SDS with NeRF representation and achieving more realistic results.

\noindent\textbf{DMTet. } 
Magic3D~\cite{magic3d22} introduces a two-stage optimization method to overcome the lack of fine details in NeRF representations. Starting with a coarse NeRF representation generated by DreamFusion~\cite{dreamfusion22}, a subsequent fine-stage optimization is performed to convert the NeRF model into a DMTet representation. This approach leverages rasterization instead of volume rendering, resulting in reduced memory requirements and increased resolution of the generated 3D objects. Fantasia3D~\cite{chen2023fantasia3d} takes a different approach by training DMTet from scratch, achieving high-quality geometry and textures. The optimization process in Fantasia3D disentangles geometry and appearance modeling using the SDS loss. To ensure realistic rendering, Fantasia3D introduces a spatially varying Bidirectional Reflectance Distribution Function (BRDF) for appearance modeling. However, a limitation of Fantasia3D is that it tends to produce materials entangled with environmental lights, neglecting specular terms in the BRDF. To address this, a subsequent work~\cite{xu2023matlaber} incorporates both diffuse and specular terms into the appearance modeling of Fantasia3D, utilizing a pretrained latent BRDF auto-encoder to ensure realistic and coherent object materials. Sherpa3D~\cite{liu2023sherpa3d} improves multi-view consistency by initializing DMTet with a coarse 3D prior from a pretrained 3D generator~\cite{jun2023shape}. Additionally, it introduces structure regularization and CLIP semantic constraints to preserve salient geometric and semantic perception from the coarse 3D prior.

\begin{figure}[t]
\centering
\includegraphics[width=1\linewidth]{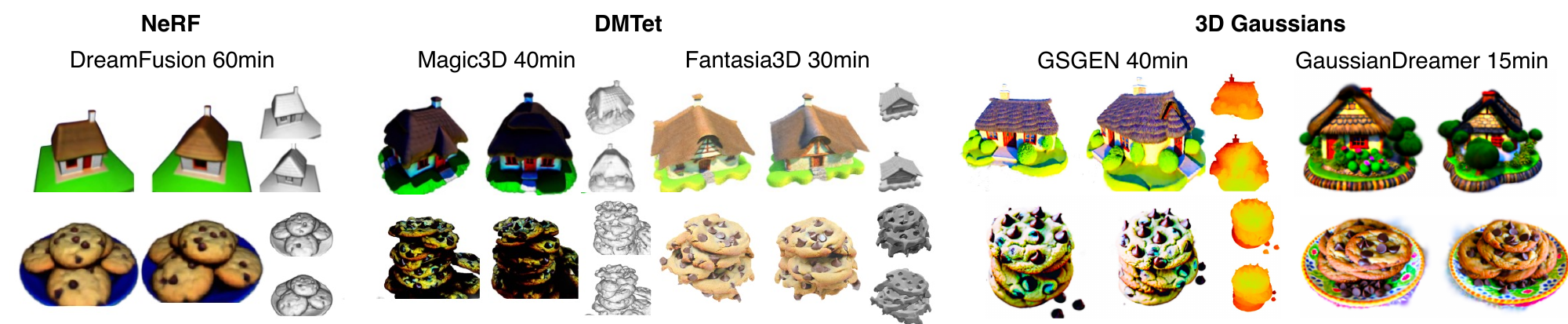}
\caption{Quality comparison of different 3d representations for optimization-based generative methods. 3D Gaussian representation shows more details and faster inference, but it is more over-saturated than DMTet. }
\label{fig-opt3drep}
\end{figure}
\noindent\textbf{3D Gaussians. } 
Another enhanced representation is the use of 3D Gaussians, which offers advantages such as reduced rendering cost and faster convergence compared to NeRF. DreamGaussian~\cite{tang2023dreamgaussian} employs a sampling approach within a sphere and optimizes 3D Gaussians using the SDS loss. The method periodically densifies points to add detail and extracts a mesh from the 3D Gaussians by locally querying density and refining UV-space to enhance texture details. However, DreamGaussian faces a significant challenge known as the Janus problem, resulting from the ambiguities of 2D SDS supervision and fast convergence. To address this problem and generate more coherent geometry, GSGEN~\cite{chen2023text} and GaussianDreamer~\cite{yi2023gaussiandreamer} introduce a coarse 3D prior. GaussianDreamer~\cite{yi2023gaussiandreamer} initializes from Shap-E~\cite{jun2023shape}, while GSGEN~\cite{chen2023text} initializes from Point-E~\cite{nichol2022pointe}. GSGEN~\cite{chen2023text} further introduces 3D SDS loss from Point-E to jointly optimize in the geometry stage. In the appearance refinement stage, GSGEN proposes a compactness-based densification strategy, filling the holes among Gaussians and their neighbors, resulting in a more complete geometry structure. Furthermore, LucidDreamer~\cite{EnVision2023luciddreamer} also utilizes 3D Gaussian representation and applies deterministic diffusing trajectories and interval-based score matching to achieve high-fidelity results.

\noindent\textbf{Disscussion. }
DMTet exhibits superior rendering speed compared to NeRF and seamlessly leverages physical rendering engines to produce realistic textures. This makes DMTet well-suited for industry applications and interactive experiences. But DMTet lacks flexibility and requires careful initialization.
On the other hand, 3D Gaussians offer faster convergence compared to NeRF and DMTet through progressive densification. Different from the DMTet representation, which needs to be initialized from tetrahedron mesh, Gaussians can initialize from point clouds, which can be obtained from trained feedforward generation methods. However, training 3D Gaussians requires hyperparameter tuning. We show the quality comparison in~\cref{fig-opt3drep}. 
In summary, DMTet and 3D Gaussians provide different benefits and trade-offs. These approaches contribute to improving the quality and fidelity of 3D representations, enabling more realistic results in various applications.

\subsection{Improved optimization strategy}\label{sec-optop}
Several methods have proposed enhanced optimization strategies to improve the fidelity of 3D content generation in text-to-3D pipelines. In particular, Magic3D~\cite{magic3d22} introduces a coarse-to-fine optimization strategy, while other approaches~\cite{huang2023dreamtime,zhu2023hifa,wang2023prolificdreamer} adopt non-increasing time sampling strategies instead of uniform time sampling. These strategies involve using large time steps to capture the global structure and gradually decreasing the time steps to capture more visual details during training iterations. DreamTime~\cite{huang2023dreamtime} further proposes a weighted non-increasing time sampling strategy. HiFA~\cite{zhu2023hifa} incorporates a denoised latent vector residual as the score instead of using noise in the optimization equation. Additionally, HiFA proposes a z-variance regularization ensuring geometrical consistency and eliminating cloudy geometrical artifacts .

To address the Janus problem in SDS, several methodss~\cite{armandpour2023re,hong2024debiasing,seo2023let} have been developed. Prep-neg~\cite{armandpour2023re} replaces simple view prompts with a combination of positive and negative prompts, limiting the direction of denoising in the DDIM. 3DFuse~\cite{seo2023let} improves 3D shape consistency by employing a consistency injection module that constructs a 3D point cloud from the text prompt and feeds its projected depth map at a given view as a condition for the diffusion model.

In addition, Latent-NeRF~\cite{metzer2023latent} accelerates the SDS optimization process by utilizing a feature space NeRF instead of an image space NeRF. It also incorporates a coarse 3D prior with a soft occupancy constraint to provide shape guidance.
DreamPropeller~\cite{zhou2023dreampropeller} leverages parallel computing techniques to achieve significant speedup, up to 4.7 times faster, with a negligible drop in generation quality. It treats parameter update rules as Picard iterations, enabling parallel sampling of an ODE path.

Overall, these improved optimization strategies contribute to more efficient and high-quality text-to-3D contents generation. Some of them, such as coarse-to-fine optimization and non-increasing time sampling, are widely used in the current optimization-based pipelines.

\subsection{Modifying SDS objective}\label{sec-optsds}
\noindent\textbf{Variational Score Distillation (VSD). }
Despite success in text-to-3D NeRF generation, SDS often suffers from over-saturation and simplistic geometry. 
To address these limitations and enhance the generative quality, ProlificDreamer~\cite{wang2023prolificdreamer} introduces a method called Variational Score Distillation (VSD). In VSD, the noise sample $\epsilon$ in~\cref{eq-SDS} is replaced with a trainable LoRA diffusion, parameterized by $\psi$, and an additional camera parameter $c$ is incorporated into the condition embeddings in the network:
\begin{equation}
    \nabla_{\theta}L_{VSD}(\theta) \triangleq \mathrm{E}_{t,\epsilon}[w(t)(\hat{\epsilon}_{\Phi}(\mathbf{x}_t, t, y)-\epsilon_\psi(\mathbf{x}_t,t,c,y))\frac{\delta g(\theta, c)}{\delta \theta}],
\end{equation}
Compared to SDS, VSD improves the diversity and sample quality by introducing a common CFG weight (e.g., 7.5). However, it is worth noting that training LoRA can be computationally expensive, as it requires double optimization compared to SDS.

\noindent\textbf{Classifier Score Distillation (CSD). }
The classifier-free guidance (CFG) is a vital and well-known trick for text-to-3D generation. When using CFG in~\cref{eq-cfg}, the gradient that drives the optimization actually comprises two terms. The primary one is estimated by the diffusion models to help move the synthesized images $x$ to high-density data regions conditioned on a text prompt $y$, which is the original optimization objective (see~\cref{eq-SDS})). The second term can be empirically interpreted as an implicit classification model. Yu et al.~\cite{yu2023text} investigate the principle of CFG and find the classifier component in~\cref{eq-cfg} is sufficient for text-to-3D generation. The proposed Classifier Score Distillation (CSD) can be found:
\begin{equation}
    \nabla_{\theta}L_{CSD}(\theta) \triangleq \mathrm{E}_{t,\epsilon}[w(t)(\epsilon_{\Phi}(\mathbf{x}_t, t, y)-\epsilon_{\Phi}(\mathbf{x}_t, t, \emptyset))\frac{\delta g(\theta, c)}{\delta \theta}],
\end{equation}
meaning rendered images at any noise level $t$ align closely with their respective noise-aware implicit classifiers.

\noindent\textbf{Interval Score Matching (ISM). }
The empirical observations have shown that SDS often encounters over-smoothing issue, LucidDreamer~\cite{EnVision2023luciddreamer} reveals that the mechanism behind SDS is to match the images rendered by the 3D model with the pseudo-Ground-Truth (pseudo-GT) generated by the diffusion model. However, the generated pseudo-GTs are usually inconsistent and have low visual quality, leading to over-smooth and lacking of details for 3D model. To address above issues, LucidDreamer proposes a Interval Score Matching (ISM). which matches between two interval steps in the diffusion trajectory instead of matching the pseudo-GTs with rendered images, and avoids one-step reconstruction that yields high reconstruction error. The gradient of ISM loss over $\theta$ is given by:
\begin{equation}
    \nabla_{\theta}L_{ISM}(\theta) \triangleq \mathrm{E}_{t,\epsilon}[w(t)(\hat{\epsilon}_{\Phi}(\mathbf{x}_t, t, y)-\hat{\epsilon}_{\Phi}(\mathbf{x}_s, s, \emptyset))\frac{\delta g(\theta, c)}{\delta \theta}],
\end{equation}

\begin{figure}[t]
\centering
\includegraphics[width=1\linewidth]{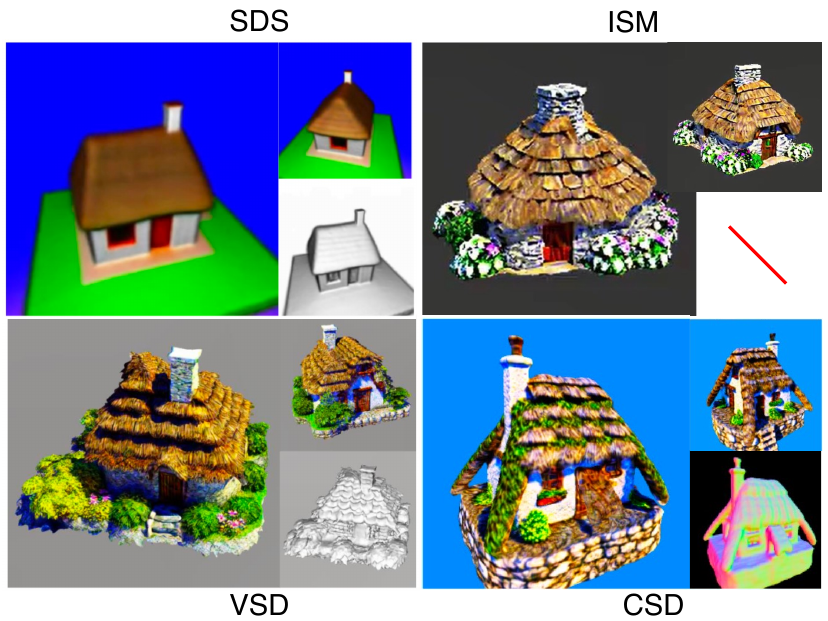}
\caption{Quality comparison of different optimization objectives for improving texture quality. The text prompt is "A 3D model of an adorable cottage with a thatched roof". }
\label{fig-optsds}
\end{figure}

\noindent\textbf{Consistency Distillation Sampling (CDS). }
The sampling process of SDS indeed corresponds to the trajectory sampling of a stochastic differential equation (SDE). However, the randomness in SDE sampling often leads to a diverse and unpredictable sample yielding inconsistent guidance in 3D generation. Inspired by the fact that an ordinary differential equation (ODE) of an SDE can provide a deterministic and consistent sampling trajectory, Consistent3D~\cite{wu2024consistent3d} introduces a Consistency Distillation Sampling loss (CDS) to enforce the optimization process of the 3D model $\Phi$ to match the deterministic flow between two adjacent ODE sampling steps.
\begin{equation}
\begin{split}
        \nabla_{\theta}L_{CDS}(\theta) \triangleq  & \mathrm{E}_{t,\epsilon}[w(t_2)(\epsilon_{\Phi}(\mathbf{x}_{t_1}, t_1, y)\\
        & -\text{sg}(\epsilon_{\Phi}(\hat{\mathbf{x}}_{t_2}, t_2, y)))\frac{\delta g(\theta, c)}{\delta \theta}],
\end{split}
\end{equation}
where $\text{sg}(\cdot)$ is a stop-gradient operator, $t_1 > t_2$ are two adjacent diffusion time steps, $\mathbf{x}_{t_1}$ is noise render image and $\hat{\mathbf{x}}_{t_2}$ a less noisy sample derived from Euler solver:
\begin{equation}
    \hat{\mathbf{x}}_{t_2}=\mathbf{x}_{t_1}+\frac{\sigma_{t_2}-\sigma_{t_1}}{\sigma_{t_1}}(\mathbf{x}_{t_1}-\epsilon_\Phi(\mathbf{x}_{t_1}, t_1, y))
\end{equation}
According to the deterministic guidance, Consistent3D alleviate multi-Janus problem in original SDS. 

\subsection{Finetuning Diffusion Prior}~\label{sec-optft}
A prevalent issue in text-to-3D shape generation is known as the Janus problem. These artifacts manifest as repeated content from different viewpoints of a 3D generation, yielding a lack of realism and coherence in the rendered views. Existing works attempt to improve the SDS optimization strategies to alleviate Janus problem, like prompt engineering~\cite{perpneg23} and initialization of coarse 3D prior~\cite{metzer2023latent,seo2023let,yi2023gaussiandreamer,chen2023text}. Alternative methods~\cite{shi2023mvdream,li2023sweetdreamer,liu2023pi3d,qiu2023richdreamer,zhao2023efficientdreamer} propose to finetune view-aware diffusion models using rendered images of 3D datasets~\cite{deitke2023objaverse,deitke2023objaversexl}.

MVDream~\cite{shi2023mvdream} addresses the problem by fine-tuning a multi-view diffusion model. In order to inherit the generalizability in 2D diffusion and to obtain the multi-view consistency in 3D datasets, they jointly fine-tune the diffusion model on real images (LAION-Aesthetics V2) and synthetic multi-view images rendered from Objaverse. EfficientDreamer~\cite{zhao2023efficientdreamer} adapts a 2D diffusion model to orthogonal-view image generation, which consists of four sub-images from orthogonal viewpoints arranged in a 2$\times$2 grid. 
Instead of relying on computationally intensive renderings and fine-tuning on both synthetic and real images in MVDream, SweetDreamer~\cite{li2023sweetdreamer} uses only low-resolution and low-cost canonical coordinates maps, and hence the coarse alignment of geometric priors is computationally efficient. PI3D~\cite{liu2023pi3d} represents a 3D shape as triplane and regards it as six pseudo-images, indicated as triplane-image. Then, PI3D finetunes a 2D diffusion model to directly output triplane images on LAION-2B with normal and depth estimation models and Objaverse. Even though triplane-image, the output of diffusion prior, saves time in SDS optimization, it needs data preprocessing before diffusion model fine-tuning, taking 2 minutes per object on a single A100 GPU. RichDreamer~\cite{qiu2023richdreamer} fine-tunes a Normal-Depth VAE and latent-diffusion model on LAION dataset together with image-to-depth and normal prior network. Then, the fine-tuned text-to-Normal-Depth diffusion model incorporates with Fantasia3D~\cite{chen2023fantasia3d} to produce 3D results.

However, the quality and diversity of fine-tuned diffusion models are far from 2D counterparts. This is partly due to the computational challenge of scaling diffusion network models up from 2D to 3D, but perhaps more so by the limited amount of available 3D training data. Some methods~\cite{shi2023mvdream,li2023sweetdreamer,li2023instant3d} observe that they are susceptible to overfitting on scarce 3D training data, compromising semantic consistency and realistic texture in text-to-3D generation.
\begin{figure}[t]
\centering
\includegraphics[width=1\linewidth]{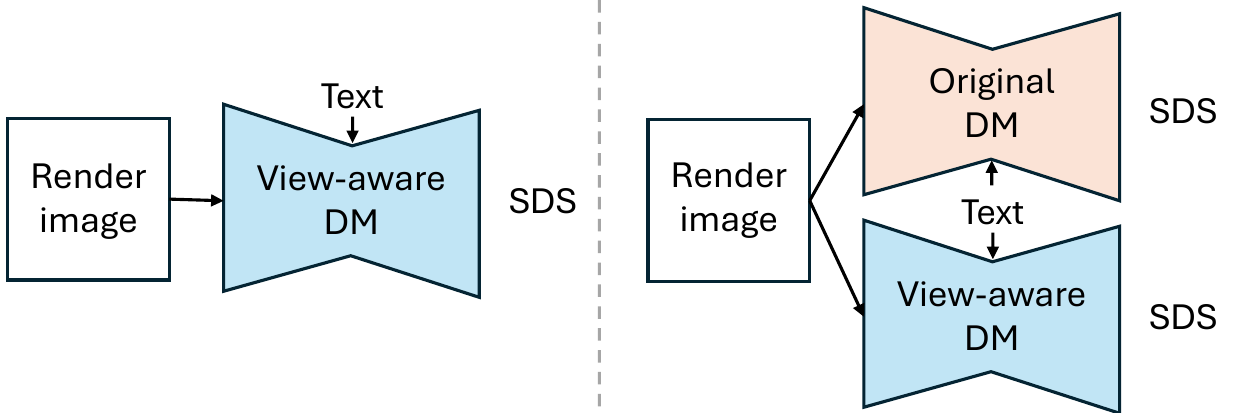}
\caption{There are two utilization for view-aware diffusion model.}
\label{fig-optft}
\end{figure}
\begin{table*}[t]
\centering
\resizebox{1\textwidth}{!}{
\begin{tabular}{c|cccccc}
\toprule
Method & Pipeline & 3D Rep. & Output View & Resolution & Data Set & Train Time (hour)\tabularnewline\midrule 
MVDream~\cite{shi2023mvdream} & Opt. & NeRF & 4 Images & 256$^2$ &  LAION (625K) + OBJ (350K) & 2304 A100 \tabularnewline
PI3D~\cite{liu2023pi3d} & Opt. & Triplane & 1 Triplane-Image & 256$^2$ & LAION (5B) + OBJ (20K) & 288 A100\tabularnewline
RichDreamer~\cite{qiu2023richdreamer} & Opt. & NeRF/DMTet & 1 Grid-Depth/Normal & 512$^2$ & LAION (2B) + OBJ (270K) & 14208 A100\tabularnewline
SweetDreamer~\cite{li2023sweetdreamer} & Opt. & NeRF/DMTet & 1 CCM & 64$^2$ & OBJ (270k) & -\tabularnewline
EfficientDreamer~\cite{zhao2023efficientdreamer} & Opt. & NeuS/DMTet & 1 2$\times$2-Grid Image & 512$^2$ & OBJ (420K) & 960 A100\tabularnewline\bottomrule
Instant3D~\cite{li2023instant3d} & MVR & Triplane & 1 Grid-Image & 1024$^2$ & OBJ (10K) & 96 A100\tabularnewline
Direct2.5~\cite{lu2023direct2} & MVR & Mesh & 1 Grid-Normal & 512$^2$ & OBJ (500K) + COYO (65M) & 3200 A100\tabularnewline
\bottomrule
\end{tabular}
}

\caption{Summary of methods finetuning text-to-image diffusion models to be view-aware, relating to optimization-based generation (Opt.) and multi-view reconstruction (MVR). The "Output View" column indicates the number and type of output views, including Canonical Coordinates Map (CCM), pseudo Triplane image, 2x2-grid image, and depth/normal map. All methods finetune on the Objaverse dataset, employing different filtering strategies to incorporate view consistency in the diffusion model. Additionally, some methods also finetune on natural images to prevent texture degradation. OBJ is the abbreviation of Objaverse.}\label{tab-dataset}
\end{table*}

\section{View Reconstruction}\label{sec:view_reconstruction}
As mentioned before, optimization-based generation requires lengthy and per-prompt optimization, hindering applicability to real-world content creation. A possible way is developing a text-to-image-to-3D approach that generates images with 2D VLMs and then trains a view reconstruction network.

\subsection{Single-View Reconstruction}
The pioneering work Point-E~\cite{nichol2022pointe} has collected millions of 3D assets along with corresponding text captions. They trained a point cloud diffusion model, which combines both the text-to-image and image-to-point cloud stages, to generate open-vocabulary 3D content. Specifically, Point-E first fine-tunes GLIDE~\cite{nichol2021glide} to generate images resembling synthetically rendered images. Then, a point cloud diffusion model is trained and conditioned on the images from the previous stage. The conditioning is achieved by leveraging the entire token sequence of the CLIP image embedding derived from the generated image. However, point cloud representation has a limited capacity to represent complex shapes. Another challenge with point cloud generation is that synthesizing high-resolution point clouds can require significant memory resources.  


\subsection{Multi-View Reconstruction}
To address the limitations of single-view reconstruction methods, several approaches have been proposed that focus on synthesizing and utilizing multiple views of the same object to improve 3D reconstruction quality.

One strategy for multi-view reconstruction involves fine-tuning 2D diffusion models to generate multiple views simultaneously. For example, Instant3D~\cite{li2023instant3d} applies this approach by fine-tuning a 2D text-to-image diffusion model to generate four-view images. These images are then used to train a large transformer-based sparse-view reconstructor to predict Triplane representation. However, the sparse number and inconsistency of views limit the reconstruction quality of Instant3D. Direct2.5~\cite{lu2023direct2} takes a different 3D representation by fine-tuning a multi-view normal diffusion model on 2.5D rendered and natural images. It produces a 2$\times$2 grid of normal maps given a text prompt. Rather than training a 3D reconstructor from multi-view images, Direct2.5 employs a fast re-meshing technique that optimizes the mesh from an initialized state using differentiable rasterization based on the produced multi-view normal maps. The optimized normal maps are then used as texture conditions to generate multi-view images.

Another approach, demonstrated by MVD$^2$\cite{zheng2024mvd}, involves directly applying a pretrained multi-view synthesis model, such as Zero123++\cite{shi2023zero123++}, to generate multi-view images given a reference image. The 3D reconstruction process in MVD$^2$ combines a pre-trained DINOv2 model and a lightweight network to predict FlexiCube representation while leveraging pixel loss supervision from normal, depth, and mask maps. Unlike MVD$^2$ represents 3D as mesh, LGM~\cite{tang2024lgm} maps multi-views into a memory-efficient 3D Gaussian representation which are produced from the pretrained MVDream~\cite{shi2023mvdream}. LGM supports higher-resolution supervision for improved reconstruction using a U-Net backbone. However, the sparse nature of the generated 3D Gaussians in LGM may limit the extraction of compact meshes. In contrast, V3D~\cite{chen2024v3d} and VideoMV~\cite{zuo2024videomv} generate dense view images using fine-tuned video diffusion models which are then converted into 3D Gaussians.

\section{Future Work}\label{sec:future_work}

The development of text-to-3D generation has advanced rapidly, but there are still a lot of challenges to overcome before they can be used for downstream applications, such as gaming, simulation, and augmented/virtual reality. Here, we discuss current gaps in the literature and potential future directions of 3D generative models.

\noindent\textbf{High-fidelity generation. }
High-fidelity generation in text-to-3D generation is an important yet under-studied area. It aims to closely align the generated 3D content with the desired attributes, shapes, features, and overall appearance specified in complex text descriptions. However, many existing text-to-3D generative methods struggle to produce fine-grained 3D content that accurately reflects the textual description, resulting in significant deviations and missing components. VP3D~\cite{chen2024vp3d} initially explores on this problem. Based on the optimization-based generative pipeline, they use the additional visual prompts as image conditions in diffusion prior, which are produced from a text-to-image diffusion model. Additionally, it introduces a differentiable reward function~\cite{xu2024imagereward} to encourage better alignment between generated 3D and 2D/text prompt. However, this method still cannot reflect complex text prompts and produces low-quality 3D shape.  A key challenge for future work is 3D generation that respects the fine-grained compositionality of the input language. We believe utilizing powerful LLM in text-to-3D diffusion models to enhance text alignment is a potential way.

\noindent\textbf{High-quality mesh. }
Mesh is the most popular 3D representation used in downstream applications. Unlike compact and topology-accurate human-designed mesh, mesh generated by iso-surface extracting techniques tends to have many sliver triangles, self-intersection, and unreasonable topology, which hamper direct application in the current graphic pipeline. One potential way is starting from template mesh and optimizing the deformation of triangles based on Neural Jacbian Field (NJF)~\cite{aigerman2022neural} following ~\cite{gao2023textdeformer}. However, NJF has a fixed triangle that limits deformation to complex shapes. A key challenge in generating high-quality mesh is developing quantity evaluation metrics.

\noindent\textbf{LLM assistant. }
LLM agents represent a new category of artificial intelligence systems built upon large models. These agents, when combined with external APIs and knowledge sources, have the potential to tackle a wide range of tasks~\cite{schick2024toolformer}. However, the application of LLM in 3D content generation is still in the exploratory phase.
One notable approach in this direction is L3GO~\cite{yamada2024l3go}, which focuses on the inference-time part-based 3D mesh generation using LLM's generate API functions within the Blender environment. However, the current capabilities of L3GO are limited to generating compositional shapes with common primitives such as cones, cubes, and cylinders. To further advance the field, we believe that LLM agents have the potential to bridge the gap between learning-based generation by networks and rule-based modeling by humans.

\noindent\textbf{Evaluation. }
Quantifying the quality of generated 3D models objectively remains a challenging and under-explored problem in the field. Existing metrics, such as CLIP Score and CLIP R-Precision, mainly focus on the association between the generated content and the text prompt. However, common metrics used in the reconstruction area, such as PSNR, SSIM, and F-Score, evaluate rendering and reconstruction results, which require ground truth data and may not provide a comprehensive assessment of the quality and diversity of the generated content.
Additionally, conducting user studies to evaluate the generated content is often time-consuming and subject to biases and the number of participants. FID captures both the quality and diversity of the results that can be applied to 3D data but may not always align with the 3D domain and human preferences and still rely on ground truth data. A recent method by Wu et al.\cite{wu2024gpt} proposes using GPT-4V to automatically generate prompts and compare generated 3D content according to user-defined criteria, which shows promise in this direction. Another approach by Ye et al.\cite{ye2024dreamreward} considers introducing human feedback as a metric and optimization objective for text-to-3D generation. Furthermore, due to the open-ended nature of the task, it is currently challenging to make fair comparisons between different methods. Therefore, the development of a comprehensive benchmark specifically designed for the text-to-3D task is of significant importance. T$^3$Bench~\cite{he2023t} has made a preliminary attempt in this direction, but further efforts are needed to establish a standardized evaluation framework that can adequately assess the performance of different text-to-3D generation methods.

\section{Conclusion}\label{sec:conclusion}

This paper presents a comprehensive review of text-to-3D generation in the wide by discussing different generative pipelines. We first introduce the fundamentals of text-to-3D generation, including formulation of large-scale Vision language models and 3D representations. Then, we review the generative pipelines that can be divided into feedforward generation, optimization-based generation and view reconstruction. Finally, we highlight the limitations of existing 3D generative models and propose several future directions. We hope that this survey will catalyze further work in text-to-3D content generation, and enable researchers to advance the state of the art. Progress in this direction has the potential to
democratize 3D content creation by enabling people to turn their
imagination into high-quality 3D assets, and to iteratively design
and control these assets for a variety of application domains.

\ifCLASSOPTIONcaptionsoff
  \newpage
\fi
\bibliographystyle{IEEEtran}
\bibliography{references}

\end{document}